\documentclass[10pt,twocolumn,letterpaper]{article}

\usepackage{cvpr}
\usepackage{times}
\usepackage{graphicx}
\usepackage{amsmath}
\usepackage{amssymb}
\usepackage{subfigure}
\usepackage{algorithmicx}

\usepackage{subfigure}

\usepackage{mathtools}
\usepackage{booktabs}
\usepackage{multirow}
\usepackage{tabu}
\usepackage{boldline}
\usepackage{algorithm}
\usepackage{algpseudocode}
\usepackage{etoolbox}
\usepackage{threeparttable}
\usepackage[font=small,labelsep=period]{caption}

\usepackage[nocompress]{cite}

\algdef{SE}[SUBALG]{Indent}{EndIndent}{}{\algorithmicend\ }%
\algtext*{Indent}
\algtext*{EndIndent}

\DeclareMathOperator*{\argmax}{arg\,max} 

\cvprfinalcopy 

\def\cvprPaperID{5380} 

\ifcvprfinal\fi
\begin{document}

\title{Efficient Neural Network Compression}

\author{Hyeji Kim, Muhammad Umar Karim Khan, and Chong-Min Kyung\\
Korea Advanced Institute of Science and Technology (KAIST), Republic of Korea\\
{\tt\small \{hyejikim89, umar, kyung\}@kaist.ac.kr}
}

\maketitle

\begin{abstract}
Network compression reduces the computational complexity and memory consumption of deep neural networks by reducing the number of parameters. In SVD-based network compression the right rank needs to be decided for every layer of the network. In this paper we propose an efficient method for obtaining the rank configuration of the whole network.
Unlike previous methods which consider each layer separately, our method considers the whole network to choose the right rank configuration. We propose novel accuracy metrics to represent the accuracy and complexity relationship for a given neural network. We use these metrics in a non-iterative fashion to obtain the right rank configuration which satisfies the constraints on FLOPs and memory while maintaining sufficient accuracy. Experiments show that our method provides better compromise between accuracy and computational complexity/memory consumption while performing compression at much higher speed. For VGG-16 our network can reduce the FLOPs by 25\% and improve accuracy by 0.7\% compared to the baseline, while requiring only 3 minutes on a CPU to search for the right rank configuration.
Previously, similar results were achieved in 4 hours with 8 GPUs. 
The proposed method can be used for lossless compression of a neural network as well. The better accuracy and complexity compromise, as well as the extremely fast speed of our method makes it suitable for neural network compression.
\end{abstract}


\section{Introduction}
Deep convolutional neural networks have been consistently showing outstanding performance in a variety of applications, however, this performance comes at a high computational cost compared to past methods. The millions of parameters of a typical neural network require immense computational power and memory for storage. Thus, model compression is required to reduce the number of parameters of the network. Our aim in this paper is to develop a method that can optimize trained neural networks for reduction in computational power and memory usage while providing competitive accuracy. 

Generally, filter pruning techniques have been used for network compression. The aim of such approaches is to develop pruning policies, which can satisfy the target constraints on FLOPs and memory. Layer wise search-based heuristic methods~\cite{cpd,fph2,fph3}, reinforcement learning~\cite{amc,fprl2,fprl3}, and genetic and evolutionary algorithms~\cite{ fpge1, fpge2} have been used to define the pruning policy. 
A greedy selection method based on a heuristic metric has been proposed in \cite{fpmulti1,fpmulti2} to prune multiple filters of the network together.

\begin{figure}[t]
\begin{center}
\includegraphics[width=0.9\linewidth]{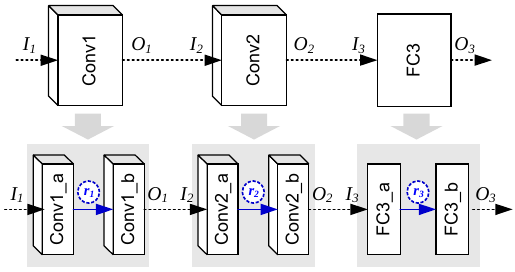}
\end{center}
\vspace{-2ex}
\caption{
Convolutional and fully-connected layers and their decomposed counterparts.
Each layer is split into two with low-rank decomposition.
}
\vspace{-4ex}
\label{fig:decomp}
\end{figure}

Another approach towards network compression is using kernel decomposition over each filter in the network. Convolutional and fully connected layers can be represented as matrix multiplications, and kernel decomposition can be applied to these matrices~\cite{dcp,dsp,ydkim,dcpd,asym}. Kernel decomposition with singular value decomposition (SVD) automatically assigns importance (the singular values) to the decomposed kernels.
This automatic sorting makes filter pruning easier, as the decomposed kernels with the lower parameters are the first to be pruned. Simply put, low-rank approximation of a layer decomposes it into two matrix multiplications for network compression as shown in Fig.~\ref{fig:decomp}.

With kernel decomposition schemes, the problem boils down to the choice of the optimal compression ratio for each layer of the network.
We need to find the right rank configuration (i.e. compression ratios) for the whole network that satisfies constraints on speed, memory and accuracy.
This is different from past methods which use a solver to minimize the approximation error at the fixed rank~\cite{ref1, lrd} and a training technique to better compensate for accuracy loss~\cite{dic, lra}.
An iterative search-based approach was adopted in \cite{asym,adc,ars} to obtain the right rank configuration.
In \cite{asym}, the authors define an PCA energy-based accuracy feature and use it to select a layer to be compressed in every iteration. The final rank of each layer is the result of iterative layer-wise network compression. 
Reinforcement learning was used in \cite{adc} to find the rank of each layer independently.
Unlike \cite{asym} and \cite{adc}, which optimize each layer separately, \cite{ars} searched for the right rank configuration for the whole network. Although, \cite{ars} shows better performance compared to \cite{asym} and \cite{adc}, it still takes significant amount of time due to its iterative search for the right rank configuration.

In this paper, we propose Efficient Neural network Compression (ENC) to obtain the optimal rank configuration for kernel decomposition. The proposed method is non-iterative; therefore, it performs compression much faster compared to numerous recent methods. Specifically, we propose three methods: ENC-Map, ENC-Model and ENC-Inf. ENC-Map uses a mapping function to obtain the right rank configuration from the given constraint on complexity. ENC-Model uses a metric representative of the accuracy of the whole network to find the right rank configuration. ENC-Inf uses both the accuracy model and inference on a validation dataset to arrive at the right rank configuration. 
The code for our method is available online.\footnote{https://github.com/Hyeji-Kim/ENC}

The rest of the paper is structured as follows. Section 2 and 3 describe the accuracy metrics. Section 4 discusses ENC-Map. Search-based methods, ENC-Model and ENC-Inf, are given in Section 5. Experiments are discussed in Section 6 and Section 7 concludes the paper.

\begin{figure}[t]
\begin{center}
\subfigure[$y_{p,l}$]{\includegraphics[width=0.494\linewidth]{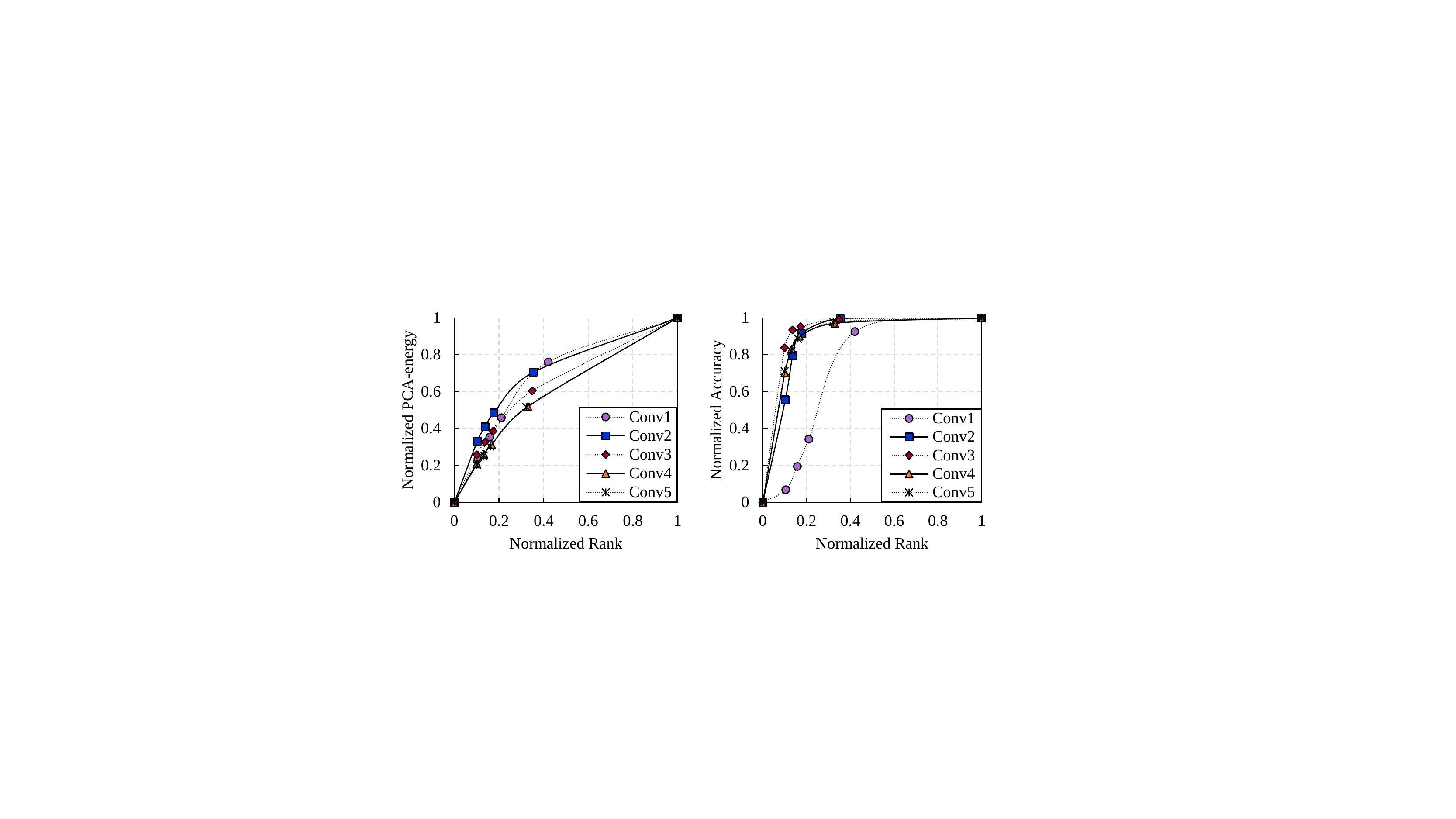}}
\subfigure[$y_{m,l}$]{\includegraphics[width=0.488\linewidth]{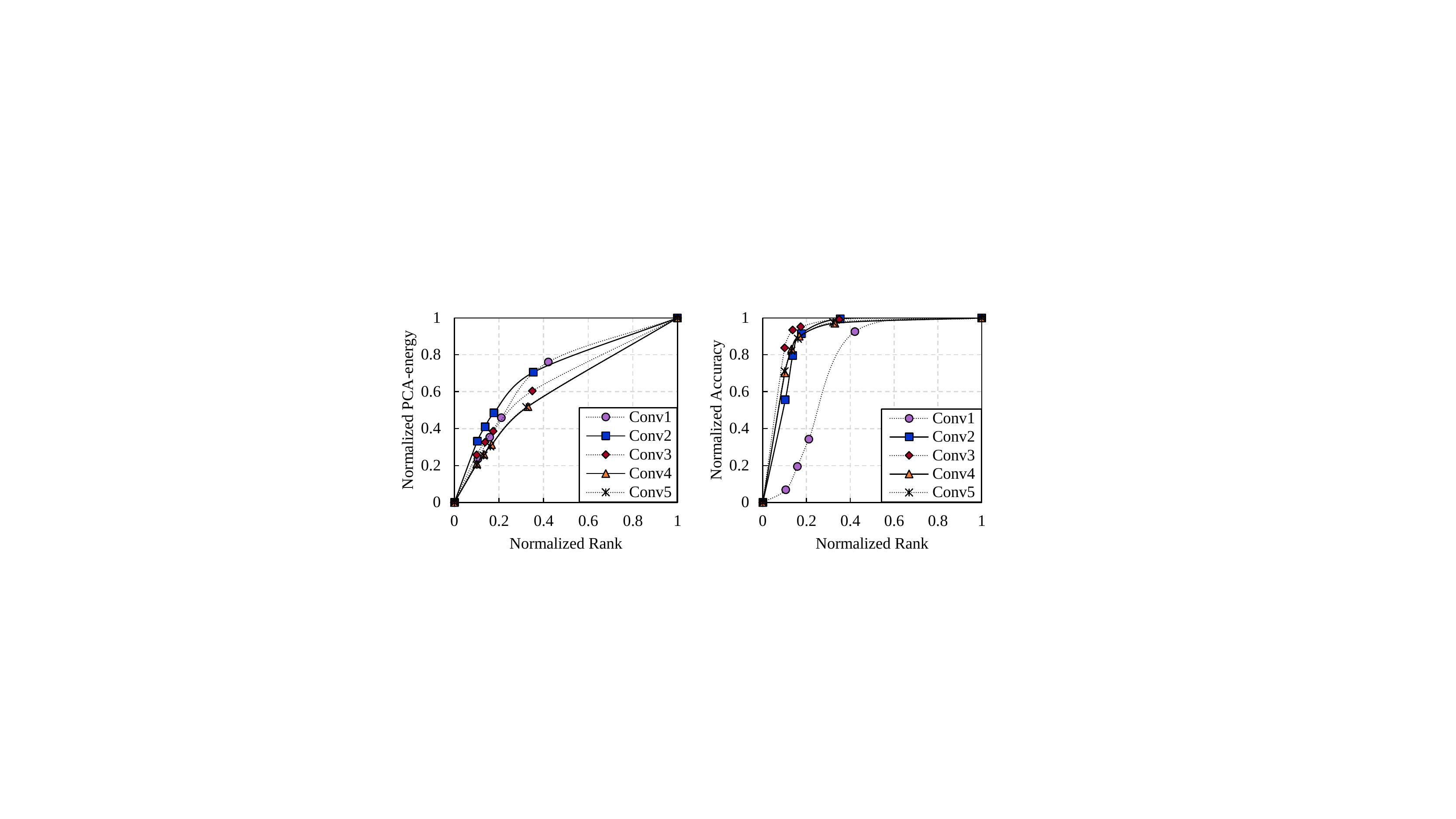}}
\end{center}
\vspace{-3ex}
\caption{Interpolated function of layer-wise accuracy metric in AlexNet. (a) $y_{p,l}$ based on normalized PCA-energy. (b) $y_{m,l}$ based on normalized accuracy using partial training dataset. }
\vspace{-3.5ex}
\label{fig:feature}
\end{figure}

\section{Layer-wise Accuracy Metrics}
The error of the neural network can be divided over the constituent layers of the network. In other words, each layer contributes to the error or accuracy of the neural network. 
In this section, we describe a couple of metrics, which represents the accuracy contributed by a layer as a function of the rank of that layer. 
This layer-wise accuracy metrics can be used to predict the contribution of an individual layer to the overall accuracy of the network. 
The first metric is based on heuristics while the second involves computing the accuracy over a validation dataset.

\textbf{PCA energy-based Metric.}\;\; 
After singular value decomposition (SVD), the number of principal components retained directly affects the complexity as well as the accuracy. The un-normalized PCA energy is given by $\sigma'_{l}(r_l)=\sum^{r_l}_{d=1}\sigma_l(d)$, 
where $r_l$ is the rank of the $l$-th layer in the neural network and $\sigma_l(d)$ is the $d$-th singular value after performing SVD on the parameters on $l$-th layer. In detail, $\sigma'_{l}(r_l)$ is the sum of the first $r_l$ diagonal entries of the diagonal matrix after decomposition. It is obvious that the accuracy decreases with the decrease in un-normalized PCA energy. The PCA energy of the $l$-th layer with a rank $r_l$ is obtained by performing min-max normalization to the un-normalized PCA energy, i.e., 
\begin{equation}\label{eq:fl_eigen}
    y_{p,l}(r_l) =
    \frac{\sigma'_{l}(r_l)-\sigma'_l(1)}{\sigma'_{l}(r_l^{max})-\sigma'_{l}(1)}.
\end{equation}
\noindent Here $r_l^{max}$ is the maximum or initial rank.

\textbf{Measurement-based Metric.}\;\; 
The second metric for layer-wise accuracy that we use in this paper is based on evaluation of the neural network on a validation dataset. For the $l$-th layer, the accuracy model is obtained by changing $r_l$ while keeping the rest of the network unchanged. Empirical models are developed for each layer. Note that the possible number of ranks for a layer occupies a large linear discrete space, making it impractical to evaluate the accuracy over the validation dataset. Therefore, we use VBMF~\cite{vbmf} to sample ranks over which the accuracy is estimated, and the ranks are sampled to increase the measured accuracy.
Then, we follow it up by Piecewise Cubic Hermite Interpolating Polynomial (PCHIP) algorithm.
We denote the measurement-based layer-wise accuracy metric of the $l$-th layer by $y_{m,l}(r_l)$.

In Fig.~\ref{fig:feature}, we show the result of interpolation with both the PCA energy-based and measurement-based approaches. Both the metrics show different profiles. This is because the PCA energy is not a representation of accuracy in itself but monotonic with the overall accuracy. 

\section{Accuracy Metric}
In this section, we describe how we can use the layer-wise metrics to represent the accuracy of the complete neural network. We estimate the joint distribution of the output of the neural network and the configuration of its layers.
The configuration of the layers of a neural network is defined only by the choice of the rank for each layer.

Our aim is to maximize the accuracy of the network with a given set of constraints. Let us define the configuration of a neural network through the rank configuration, which is a set of ranks of each layer, i.e., 
\begin{equation}\label{eq:rank_config}
R = \{r_1, r_2, ..., r_L\},
\end{equation}
\noindent where $r_l$ is the rank of the $l$-th layer. The overall accuracy of the network can be represented by the joint distribution of accuracy of individual layers. Precisely,
\begin{equation}\label{eq:overall_acc}
\textrm{P}(A; R) = \textrm{P}(a_1, a_2, ..., a_L; R),
\end{equation}
\noindent where $a_l$ is the accuracy provided by the $l$-th layer. 
Practically, the accuracy contribution of a layer also depends on the ranks of other layers. 
However, the accuracy model can be simplified as the function of layer-wise accuracy metric contribution of a layer by applying the layer-wise accuracy metrics in Sec. 2, $y_{p,l}(r_l)$ and $y_{m,l}(r_l)$, which depend on the rank of the layer only.
Considering the independence, we can model the overall accuracy metric as 
\begin{equation}\label{eq:accuracy_split}
\textrm{P}(A; R) = \prod_{l=1}^{L} \textrm{P}(a_l; r_l).
\end{equation}
\noindent This representation has been inspired by \cite{asym} where the product of layer-wise accuracy metrics is used to estimate the overall accuracy. 


To estimate the accuracy, we define three types of overall accuracy metric; measurement-based $A_m(R)$, PCA energy-based $A_p(R)$, and the combination of two metrics, $A_c(R)$. 
The layer-wise metric based on the measured accuracy can be directly used to replace $\textrm{P}(a_l; r_l)$, i.e.,
\begin{equation} \label{eqn:F_m}
     A_{m}(R)=\prod^{L}_{l=1}y_{m,l}(r_l).
\end{equation}
\noindent Although the normalized PCA energy of (\ref{eq:fl_eigen}) is not defined from the accuracy, it was shown in~\cite{asym} that the PCA energy of a layer is proportional to the accuracy of the network, allowing us to use the PCA energy as the accuracy metric. Thus, we define the PCA energy-based accuracy metric as 
\begin{equation}\label{eqn:F_p}
A_{p}(R)=\prod^{L}_{l=1} y_{p,l}(r_l).
\end{equation}
\noindent 
In our experiments, we have noticed that the PCA energy metric does not accurately represent the accuracy at very low complexity.
Also, as the network is redundant, the measurement-based metric can not sufficiently represent the effect of complexity reduction against accuracy at higher complexities.
Therefore, we define a combined metric. 
This metric takes into consideration both the PCA energy and the measurement based layer-wise accuracy metrics. 
We have weighted the PCA energy only with the network complexity to reduce the influence of the $A_p(R)$ at lower complexities and the $A_m(R)$ at higher complexities.
Mathematically,
\begin{equation}\label{eqn:F_c}
    A_c(R)=\left \{ A_p(R) \times \frac{C(R)}{C_{orig}}  \right \} + A_m(R),
\end{equation}
\noindent where $C(R)$ is the complexity of the rank configuration $R$ and $C_{orig}$ is the total complexity of the network. The complexity $C(R)$ is defined by
\begin{equation}\label{eq:tot_complexity}
    C(R) = \sum_{l=1}^{L} C_l(r_l) = \sum_{l=1}^{L} c_lr_l,
\end{equation}
\noindent where $C_l(r_l)$ is the complexity of the $l$-th layer. The complexity coefficient $c_l$ for spatial~\cite{dsp} and channel~\cite{dcp} decomposition is $c_l=W_lH_lD_l(I_l+O_l)$ and $c_l=W_lH_l(I_lD_l^2 + O_l)$, respectively.
Here $W_l$ and $H_l$ are the width and height of output feature map, $D_l$ is the size of filter window, $I_l$ and $O_l$ are the number of input channels and filters.

As will be seen shortly, we are not interested in the exact value of the estimated accuracy but rather the relative value of accuracy metric obtained from the various rank configurations.
In other words, we use the accuracy metric to extract some partial rank configurations with the largest value of accuracy metric. 

\section{ENC-Map: Rank Configuration with \\Accuracy-Complexity Mapping}\label{sec:simple_method}
In this section, we present a simple method to choose the rank configuration for a neural network by mapping complexity against accuracy. The mapping is performed through the rank configurations, as both complexity and accuracy are functions of rank configurations.
At first, we intuitively though that all layers having same accuracy penalty would be better compression strategy than having same compression ratio (i.e. uniform). 
Therefore, we only consider the rank configurations for which layer-wise metrics are equal for every layer. 
Mathematically,
\begin{equation}
    R_e =  R\;|\;y_{i, l}(r_l) = y_{i, k}(r_k), 
\end{equation}
\noindent where $l, k \in 1, 2, ..., L$ and $i \in \{p, m\}$
.
Next the complexity of $R_e$, $C(R_e)$, is computed using (\ref{eq:tot_complexity}). 
The accuracy metric and complexity are plotted against each other over $R_e$, which provides a mapping between complexity and accuracy metric. The mapping is mathematically given as
\begin{equation}\label{eq:map_c2a}
    f_{C-A}:\mathbb{R} \rightarrow \mathbb{R}.
\end{equation}
\noindent Also note that the mappings from complexity and accuracy metric to the rank configuration and vice versa do exist as well, i.e.,
\begin{equation}\label{eq:map_c2r}
    f_{C-R}:\mathbb{R} \rightarrow \mathbb{R}^{L}.
\end{equation}
\noindent
Using this mapping, the respective rank configuration is calculated from the inverse function of layer-wise accuracy metric, where $a=y_{k, l}(r_l)$ is converted to $r_l=y_{k, l}^{-1}(a)$ and $k\in \{p, m\}$. 
Note that the inverse exists as $y_{k, l}(r_l)$ is an increasing function as shown in Fig.~\ref{fig:feature}.
Here, $a$ is the achievable layer-wise accuracy metric obtained from $f_{C-A}$, while satisfying constraints on complexity.
This method is called ENC-Map as we use a simple mapping to obtain the right rank configuration.

\begin{figure}[t]
\begin{center}
\includegraphics[width=0.98\linewidth]{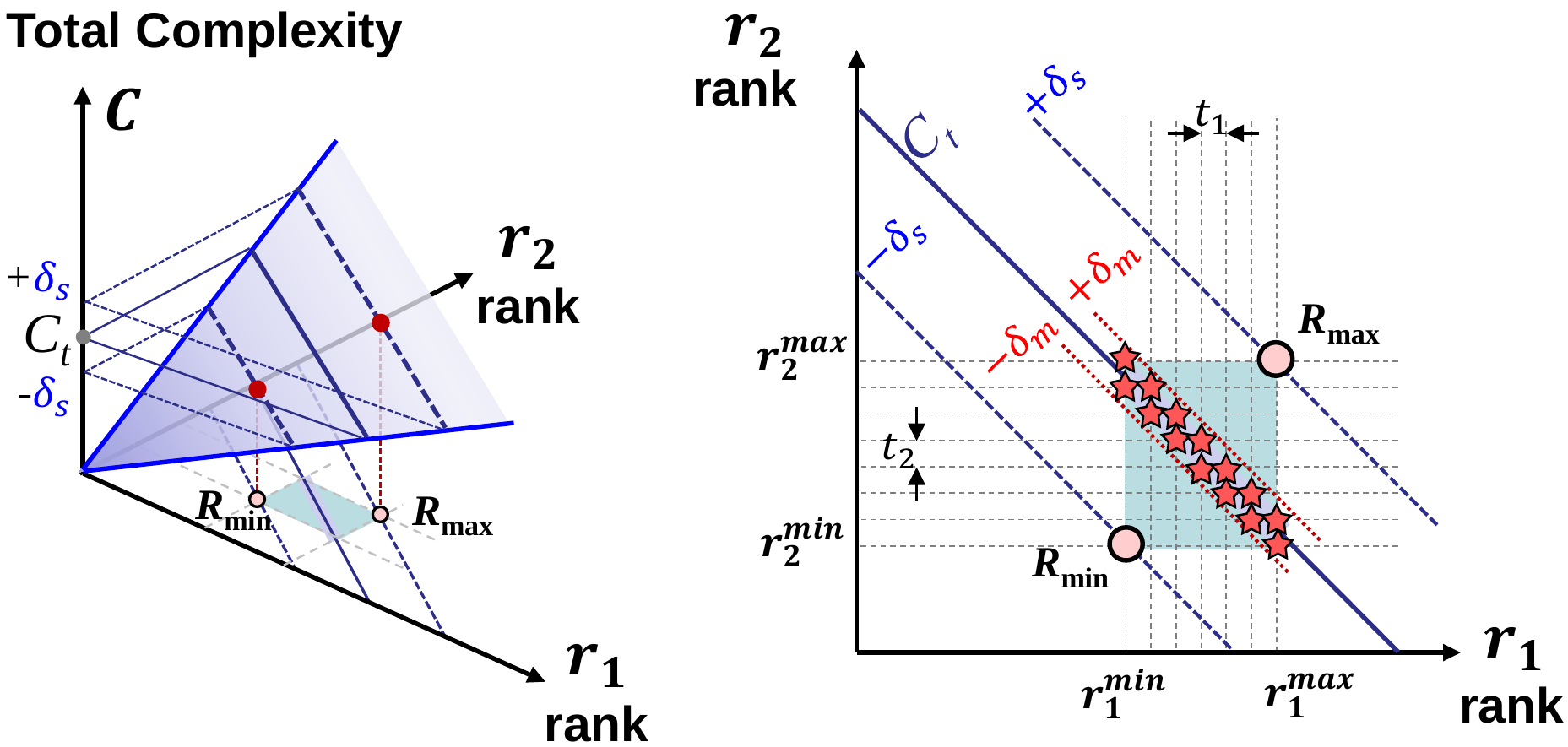}
\end{center}
\vspace{-1.0ex}
\caption{Extraction of candidate rank configurations (as an example of two-layered CNN). The effective space is defined by boundaries of rank configuration, $R_{max}$ and $R_{min}$, and step size $t_l$, where the complexity is around target complexity $C_t$ with space margin $\pm\delta_s$.
The candidate rank configurations are lying on $C_t$ with the complexity margin $\pm\delta_m$ denoted as star points.}


\vspace{-2.0ex}
\label{fig:space_gen}
\end{figure}

\section{ENC-Model/Inf: Rank Configuration in\\ Combinatorial Space}
The method described in the previous section strongly depends on the equal layer-wise metrics.
Therefore, we extend the rank configuration to the combinatorial problem to choose the optimal rank configuration in the non-equal layer-wise metrics. 


The combinatorial space is defined by the Cartesian product of the vector space of rank for each layer.
In this space, we extract the partial rank configurations that satisfy the target complexity called candidate rank configurations as illustrated in Fig.~\ref{fig:space_gen}.
To quickly extract the candidate rank configurations and find the optimal rank configuration in non-iterative manner, we perform two steps. First, we limit the range of the ranks with ENC-Map. Then, we extract the candidate rank configurations by hierarchically grouping the layers to reduce the number of effective layers for sub-space generation.



\subsection{Limiting the Search Space}
\begin{figure}[t]
\begin{center}
\includegraphics[width=0.85\linewidth]{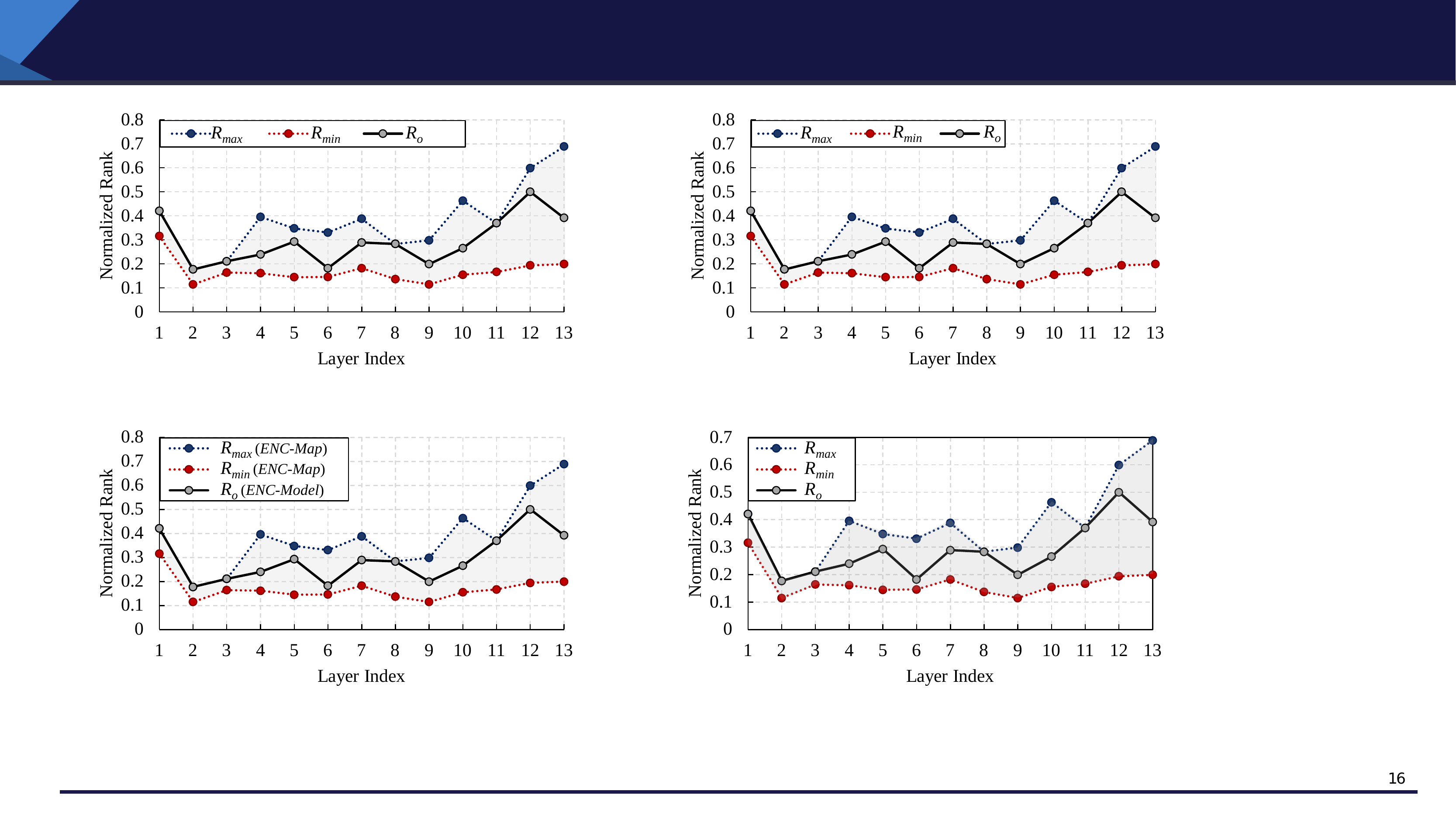}
\end{center}
\vspace{-1.0ex}
\caption{Distribution of normalized rank configurations for VGG-16 ($C_t$=25\%, $\delta_s$=$10\%$). A final rank configuration $R_o$ is selected within the boundaries, $R_{max}$($C$=35\%) and $R_{min}$($C$=15\%). }
\vspace{-2.0ex}
\label{fig:space_r}
\end{figure}

To limit the search space, we use the simple method described in the previous section. We obtain the upper and lower bounds on the search space near the target complexity by using the mapping in (\ref{eq:map_c2r}) by
\begin{equation}\label{eq:upperbound}
    R_{max} = f_{C-R}(C_t + \delta_s), 
\end{equation}
\begin{equation}\label{eq:upperbound}
    R_{min} = f_{C-R}(C_t - \delta_s).
\end{equation}
\noindent Here $C_t$ is the complexity constraint and $\delta_s$ is the space margin.
The final rank configuration, $R_o$, is defined within the space boundaries, $R_{max}$ and $R_{min}$, as shown in Fig.~\ref{fig:space_r}.


\subsection{Min-offsetting the Search Space}
In the limited space by the $R_{max}$ and $R_{min}$, we are only interested in the candidate rank configurations $\mathbf{R}$ as
\begin{equation}\label{eqn:candidate}
    \mathbf{R} = \bigcup R\;|\;C_t-\delta_m <C(R)<C_t+\delta_m,
\end{equation}
for $R\in [R_{min},R_{max}]$.
$\mathbf{R}$ includes the rank configurations for which the complexity is roughly equal to $C_t$. Note that we slightly expand the search space by using the parameter $\delta_m$.
To simplify the computations, we shift the space boundaries from $[R_{min},R_{max}]$ to $[0,R_{max}-R_{min}]$.
Then, we generate the differential search space and extract the differential candidate rank configurations $\hat{\mathbf{R}}$ defined by  
\begin{equation}\label{eqn:candidate_}
    \hat{\mathbf{R}} = \bigcup R\;|\;\Delta C_t-\delta_m <C(R)<\Delta C_t+\delta_m,
\end{equation}
for $\Delta C_t=C(R_{max})-C_t$ and $R\in [0,R_{max}-R_{min}]$.
Note that $\mathbf{R} = R_{max} - \hat{\mathbf{R}}$.

\subsection{Hierarchical Extraction of $\hat{\mathbf{R}}$}
\begin{figure}[t]
\begin{center}
\includegraphics[width=0.85\linewidth]{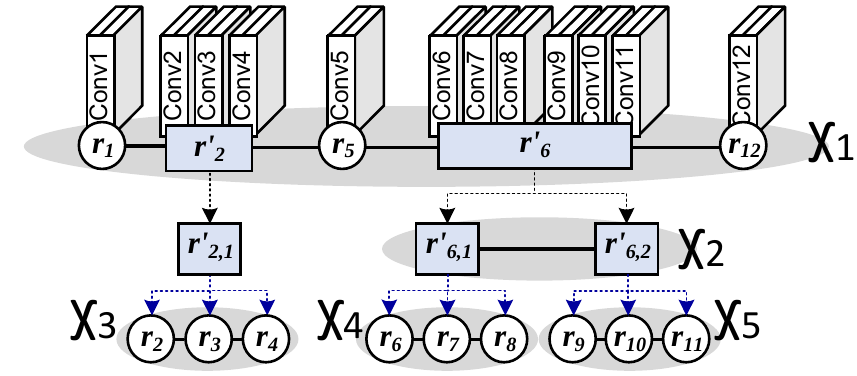}
\end{center}
\vspace{-1.0ex}
\caption{Hierarchical space generation (as an example of 12-layered network). There are 3 hierarchy level and five sub-spaces ($\mathcal{X}_1,...,\mathcal{X}_5$). The maximum space complexity is $\mathcal{O}(n^5)$ of $\mathcal{X}_1$.}
\vspace{-3.0ex}
\label{fig:group}
\end{figure}

It is not feasible to generate the complete combinatorial search space for deep neural network, since the space complexity is exponential to the number of layers.
Hence, we hierarchically generate the sub-spaces by grouping some layers in a top-down manner as illustrated in Fig.~\ref{fig:group} and extract the candidate rank configurations. 

As denoted in~(\ref{eq:tot_complexity}), the complexity is the weighted sum of the complexity coefficient and rank of each layer.
From the min-offset space, layers having same complexity coefficient $c_i$ can simply be grouped from $\{r_i, r_{i+1}, r_{i+2}, ...\}$ to $r'_i$.  
The groped rank $r'_i$ is defined by
\begin{equation}\label{eqn:grouped_rank}
    r'_i=[0:\min(\{t_i\}):\sum \{max(r_i)\}],
\end{equation}
where $i$ is the index of first layer in a group. 
Here, the range of rank in min-offset space is represented with the vector space and the $t_i$ means the step size of vector space.

As an example, a 12-layered CNN is illustrated in Fig.~\ref{fig:group}.
The maximum number of layers in a bottom group is set to 3, so that there are 3-hierarchical levels for layer grouping. The maximum space complexity is reduced from $\mathcal{O}(n^{12})$ to $\mathcal{O}(n^5)$ in $\mathcal{X}_1$. 
At the top-level space, $\mathcal{X}_1$ is composed of 5 vector spaces, and we only retain the rank configurations that satisfy $C_t$ in $\mathcal{X}_1$. 
The sub-space $\mathcal{X}_2$ is defined by two vector spaces, which are also the grouped variables from the bottom layers in $\mathcal{X}_4$ and $\mathcal{X}_5$.
To simplify the extraction strategy, in our experiments, we choose the partial sets in each bottom sub-spaces $(\mathcal{X}_3, \mathcal{X}_4, \mathcal{X}_5)$, by measuring the accuracy metric of each rank configuration.







\subsection{Choice of Rank Configuration}
Apart from the simple method described in Sec. \ref{sec:simple_method}, 
we use two methods, ENC-Model and ENC-Inf, for obtaining the optimal rank configuration for a given neural network.
For both the methods, we prepare a subset $\mathbf{R}$ using (\ref{eqn:candidate}) including the candidate rank configurations.



In ENC-Model, we choose a rank configuration $R_o$ that maximizes $A(R)$ which can be one of $A_m(R)$, $A_p(R)$, and $A_c(R)$ denoted in (\ref{eqn:F_m}, \ref{eqn:F_p}, \ref{eqn:F_c}). I.e.,
\begin{equation}
    R_o = \argmax_R A(R) | R \in \mathbf{R}. 
\end{equation}

In ENC-Inf, we choose $N$ rank configurations, which provide the largest values of $A(R)$. These $N$ rank configurations are stored in $\mathbf{R_A}$ and then evaluated over the validation dataset to choose a best rank configuration.

The complete process is given in Algorithm~\ref{algo:process}. The ENC-Map provides a quick and dirty solution, whereas ENC-Model and ENC-Inf approaches are relatively slower. However, even our slowest method easily outperforms state-of-the-art approaches in speed.

The proposed method can be used to reduce both the FLOPs and the memory consumption of a neural network. The constraint $C_t$ can represent both the number of FLOPs or the number of parameters of the neural network.
Also, our proposed methods can provide rank configurations under both complexity and accuracy constraints. Till now, the discussion followed complexity constrained systems. In other words, given a complexity constraint our aim is to obtain a rank configuration that maximizes the accuracy. However, shifting to accuracy constrained systems is straightforward. Given an accuracy constraint, we use the inverse mapping of $f_{C-A}$, $f_{A-C}$, to obtain the complexity corresponding to a given accuracy. The obtained complexity is input to Algorithm~\ref{algo:process} to obtain the rank configuration.

\begin{algorithm} [t]
\caption{: Optimal Rank Configuration} 
    \begin{algorithmic}
        \State
        \textbf{INPUTS}: $h \leftarrow$  {A neural network}, $C_t \leftarrow$ {Target complexity},
        $method \leftarrow $ {ENC-Map, ENC-Model, ENC-Inf}
        \State \textbf{OUTPUT}: $R_o \leftarrow$ {Rank configuration}
        \State \textbf{Parameters}: 
        $\delta_s \leftarrow$ {Parameter for limiting the search space},
        $\delta_m \leftarrow$ {Parameter for marginal error of $C_t$},
        $N \leftarrow$ {Number of rank configurations for evaluation}
        \State//{\texttt{Layer-wise metrics}}
        \State {Compute } $y_{p, l}$ , $y_{m, l}$ {for all} $l \in 1...L$ 
        \State {Compute } $f_{C-A}$, $f_{C-R}$
        \If {$method$ == {ENC-Map}}
            \State $R_o = f_{C-R}(C_t)$
        \Else
            \State //{\texttt{Candidate rank-set generation}} 
            \State $R_{max} = f_{C-R}(C_t + \delta_s)$ 
            \State $R_{min} = f_{C-R}(C_t - \delta_s)$
            \State $\mathbf{R}=\{\cup R\;|\;C_t-\delta_m <C(R)<C_t+\delta_m\}$
            \State {\;\;\;\;\;\;\;\;\;\; for $R \in [R_{min}, R_{max}]$}
            \State //{\texttt{Accuracy estimate for the}} 
            \State //{\texttt{\;whole neural network}}
            \State {Compute} $A$ of $\mathbf{R}$
            \If{$method$ == {ENC-Model}}
                \State $R_o = \argmax_{R} A(R)$ s.t. $R \in [R_{min}, R_{max}]$
            \ElsIf{$method$ == ENC-Inf}
                \State {Compute} $\mathbf{R_A} \in \mathbb{R}^{N \times L}$ 
                \State {Evaluate accuracy of each row of $\mathbf{R_A}$ over}
                \State {\;\;the validation dataset}
                \State {Select $R_o$ from $\mathbf{R_A}$ with best evaluation}
            \EndIf 
        \EndIf
    \end{algorithmic}
    \label{algo:process}
\end{algorithm}

\begin{figure*}[t]
\begin{center}
\subfigure[AlexNet]{\includegraphics[width=0.33\linewidth]{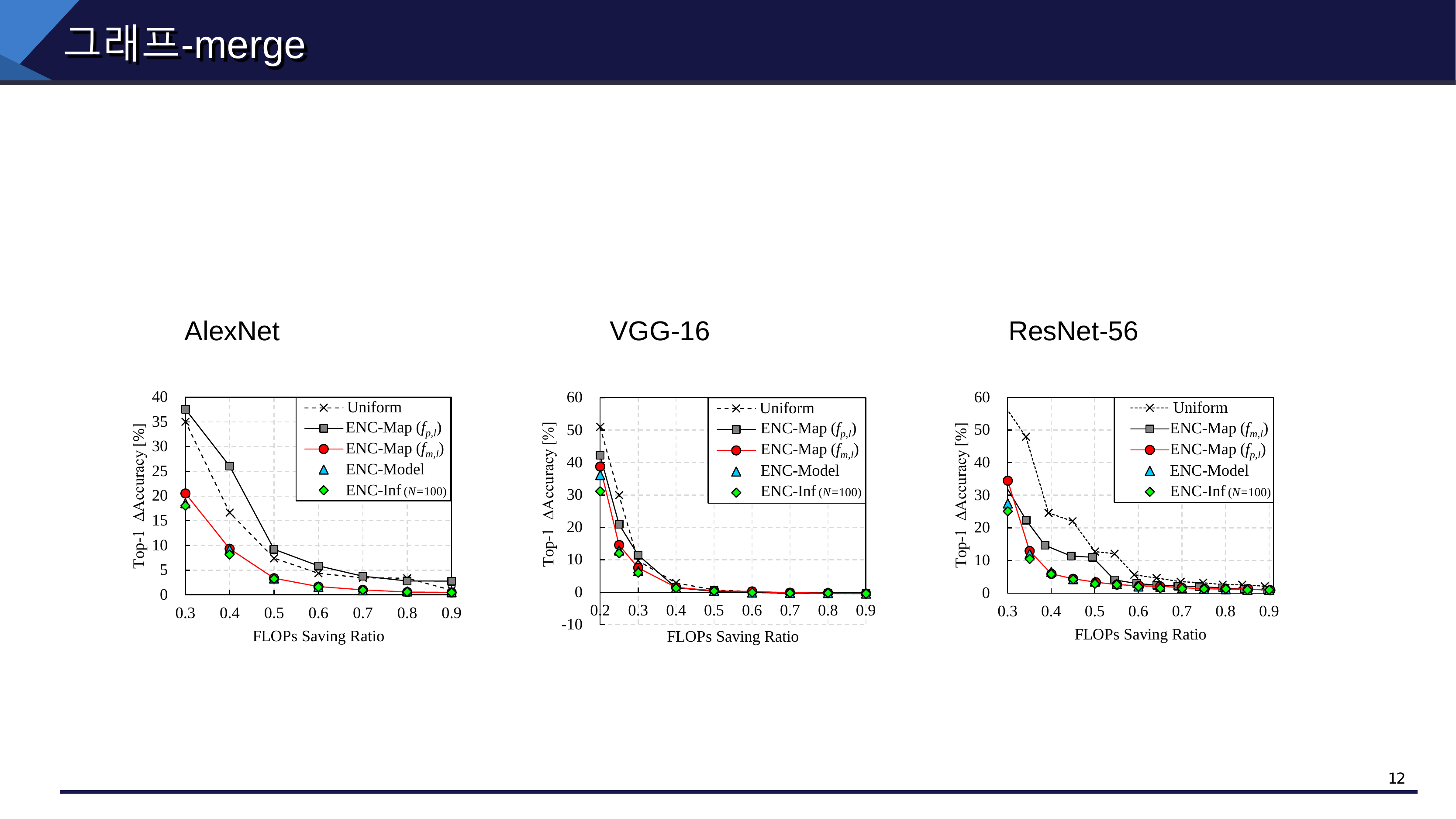}}
\subfigure[VGG-16]{\includegraphics[width=0.335\linewidth]{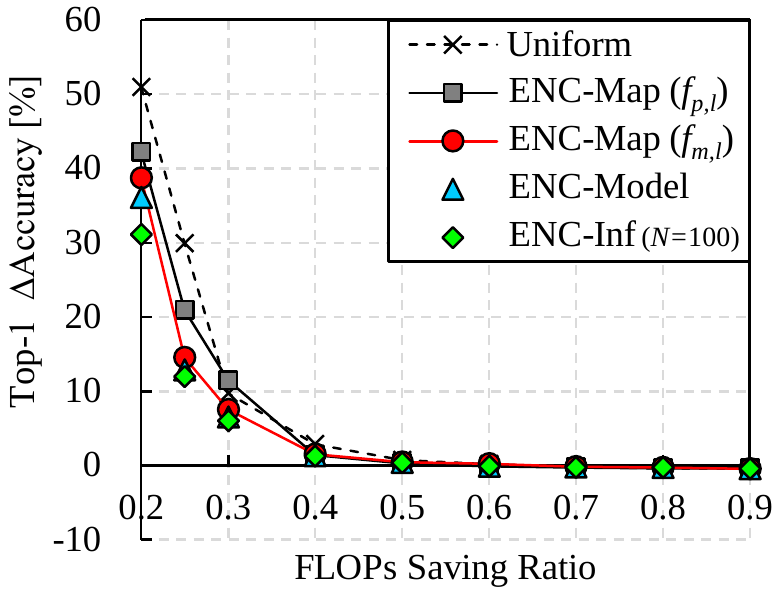}}
\subfigure[ResNet-56]{\includegraphics[width=0.32\linewidth]{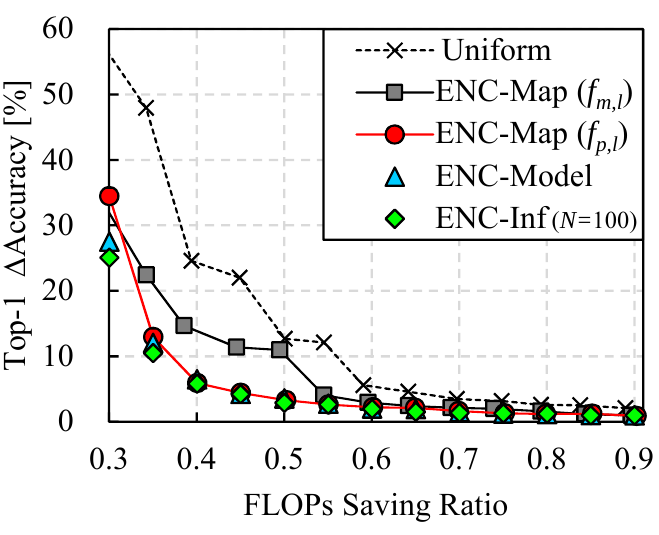}}
\end{center}
\vspace{-2.5ex}
\caption{
Performance comparison of overall accuracy metrics. Baseline top-1 accuracy is 56.6\% for AlexNet, 70.6\% for VGG-16, and 93.1\% for ResNet-56. The results are without fine-tuning. Smaller $\Delta$ Acc. is better.}
\vspace{-3ex}
\label{fig:compare_all}
\end{figure*}


\section{Experimental Results}
In this section, we present the experimental evaluation of the proposed methods. We first present a comparison of the proposed methods. We discuss the scenarios that dictate the choice of any of the proposed methods. Afterwards, we compare the proposed methods against some recently proposed neural-network optimizing methods.

For our experiments, we have optimized AlexNet~\cite{alex}, VGG-16~\cite{vgg16}  and ResNet-56~\cite{res56} networks. For AlexNet and VGG-16, we have used the ImageNet \cite{imagenet} dataset. For ResNet-56, we have used the Cifar-10 \cite{cifar10} dataset. We have performed the same pre-processing (image cropping and resizing) as described in the original papers \cite{alex,vgg16,res56}. Also, we have used 4\% and 10\% of the ImageNet and Cifar-10 training datasets as the validation datasets. 

The parameters $\delta_s$ and $\delta_m$ were set to 10\% of the original total complexity and 0.5\% of the target complexity, respectively. 
For ENC-Model and ENC-Inf, we use the accuracy metric $A_m(R)$ for AlexNet, $A_c(R)$ for VGG-16, and $A_p(R)$ for ResNet-56 as $A(R)$.
Also, note that we have performed channel decomposition to the first layer whenever required and spatial decomposition to the rest of the layers using truncated SVD. For VGG-16, the first layer was decomposed to half the original rank following~\cite{ars} and for ResNet-56, the first layer was not compressed at all.

The experiments were conducted on a system with four Nvidia GTX 1080ti GPUs and an Intel Zeon E5-2620 CPU. The experiments conducted on the CPU used a single core and code optimization was not performed. The \textit{Caffe}~\cite{caffe} library was used for development.

\begin{table*}[t]
\centering
\begin{tabular}{cccccc}
\toprule
\multirow{2}{*}{Model}& 
\multirow{2}{*}{\begin{tabular}[c]{@{}c@{}}Target\\ Complexity\end{tabular}}& \multirow{2}{*}{\begin{tabular}[c]{@{}c@{}}Searching\\ Policy\end{tabular}} & 
\multicolumn{2}{c}{Top-1/Top-5 $\Delta$Acc.[\%]} & 
\multirow{2}{*}{\begin{tabular}[c]{@{}c@{}}Search\\ Time\end{tabular}} \\ \cmidrule(l){4-5}
&&&  \begin{tabular}[c]{@{}c@{}}w/o FT\end{tabular} & \begin{tabular}[c]{@{}c@{}}w/ FT\end{tabular} &\\ \midrule

\multirow{3}{*}{\begin{tabular}[c]{@{}c@{}}AlexNet\\ (56.6\% / 79.9\%)\\ @ImageNet\end{tabular}} & \multirow{3}{*}{\begin{tabular}[c]{@{}c@{}}FLOPs 37.5\%\\ Parameters 18.4\%\\ \cite{ydkim}\end{tabular}} 
& Y-D \textit{et al}.~\cite{ydkim}& - / -& - / 1.6& -\\
&& \textbf{ENC-Inf}& - / -& -\textbf{0.1} / -\textbf{0.2}&\textbf{ 3m @4GPUs}\\
&& {ENC-Model}& - / -& {-0.1} / {-0.2}& {1m @CPU}\\
 \midrule

\multirow{8}{*}{\begin{tabular}[c]{@{}c@{}}VGG-16\\ (70.6\% / 89.9\%)\\ @ImageNet\end{tabular}} & \multirow{8}{*}{FLOPs 25\%}
& Uniform& 29.9 / 23.6&  0.8 / 0.5& -\\
&& Heuristic~\cite{adc}&- / 11.7& - / -& -\\
&& ADC~\cite{adc}&- / 9.2& - / -& 4h @8GPUs\\
&& ARS~\cite{ars}& 13.0 / 9.1& -0.3 / -0.1& \begin{tabular}[c]{@{}c@{}}7h @4GPUs \vspace{-0.8ex}\\ (in our impl.)\end{tabular}             \\
&& \textbf{ENC-Inf}&\textbf{ 10.7} / \textbf{7.3}&\textbf{ -0.6} / \textbf{-0.2}& \textbf{5m @4GPUs}\\
&& ENC-Model& 11.3 / 7.9& -0.7 / -0.2& 3m @CPU\\
&& ENC-Map & 14.5 / 10.2& -0.2 / -0.1& 4s @CPU\\ \midrule

\multirow{5}{*}{\begin{tabular}[c]{@{}c@{}}ResNet-56\\ (93.1\% / -)\\ @Cifar-10\end{tabular}}& \multirow{5}{*}{FLOPs 50\%}
& Uniform& 12.7 / -& 0.2 / -& -\\
&& AMC~\cite{amc}$^*$ & 2.7$^*$ / -& 0.9$^*$ / -& 1h @GPU\\
&&\textbf{ ENC-Inf}& \textbf{2.9} / -&\textbf{ 0.1} / -& \textbf{3m @4GPUs}\\
&& ENC-Model& 3.5 / -& 0.1 / -& 1m @CPU\\
&& ENC-Map& 3.3 / -& 0.1 / -& 3s @CPU

\\ \bottomrule
\end{tabular}
\caption{Performance comparison of network compression techniques. The proposed methods, uniform, \cite{adc}, heuristic in \cite{adc}, and \cite{ars} use the SVD based spatial decomposition. \cite{ydkim} uses the Tucker decomposition based channel decomposition. \cite{amc} uses the channel pruning. Baseline accuracy of~\cite{amc}$^*$ is 92.8\%. Smaller $\Delta$Acc. is better.}
\vspace{-2ex}
\label{table:cost_all}
\end{table*}

\subsection{Comparison of Layer-wise accuracy metrics}
This section presents the comparison of the two layer-wise accuracy metrics based on the results of the ENC-Map as shown in Fig. \ref{fig:compare_all}. 
We note that the PCA energy-based metric shows poor performance with the shallower AlexNet. This is expected as the PCA energy is not based on actual inference and expected to show poorer performance compared to the measurement-based layer-wise accuracy metric. However, with the deeper VGG-16 and ResNet-56, the performance of PCA energy-based and the measurement-based layer-wise accuracy metric is almost same. 
The reason is that the complexity difference from rank reduction on each layer is less as the layer is deeper, so that the accuracy difference is also smaller.
The measurement-based metric can not completely represent the overall accuracy.


The PCA energy-based and measurement-based metrics require 5 seconds and 5 minutes for ResNet-56. Note that once these metrics are evaluated, they can be used for different complexity constraints. For every new neural network that needs to be optimized, these metrics need to evaluated beforehand. 

\subsection{Comparison of the Fast, Search without Inference, and Search with Inference Methods}
In this section, we compare the performance of ENC-Map, ENC-Model and ENC-Inf against each other.
In Fig.~\ref{fig:compare_all}, it is seen that at lower compression (or relatively higher complexity requirements), the performance of all the three methods is almost the same.
It means that the ENC-Map is the best solution at lower compression.
At higher compression, the optimal rank configuration is more sensitive to the complexity and accuracy metric. Therefore, the performance of ENC-Map is lower than ENC-Model and ENC-Inf, and ENC-Inf using validation accuracy shows relatively the best performance.
The time taken by each model is given in Table \ref{table:cost_all}.

\subsection{Performance Comparison with Other Methods}
The overall results are summarized in Table~\ref{table:cost_all}. Here we use the decrease in accuracy (i.e. $\Delta$Acc.), which is the difference of accuracy of the original and optimized neural networks, as a metric to evaluate different methods. Lower decrease in accuracy is desired. 
For a complexity metric, we use the percentage of ($1-$compression ratio) as FLOPs.
Also, we mention the results with uniform rank reduction, which applies the same rank reduction ratio to every layer of the neural network.

\begin{figure}[t]
\begin{center}
\vspace{-2.5ex}
\subfigure[]{\includegraphics[width=0.49\linewidth]{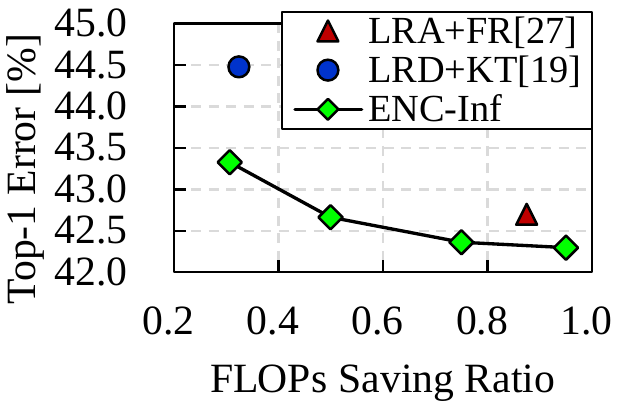}}
\subfigure[]{\includegraphics[width=0.49\linewidth]{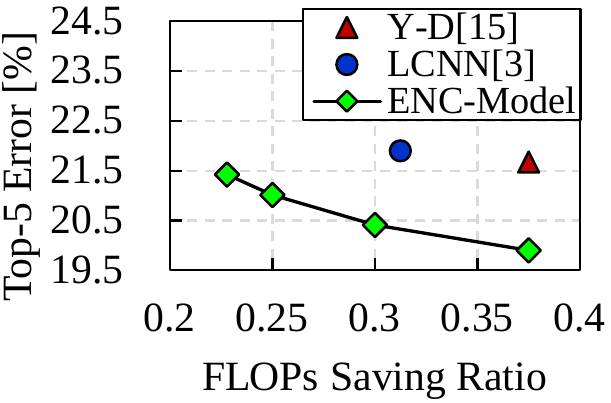}}
\end{center}
\vspace{-4.5ex}
\caption{Performance comparison for AlexNet. (a) compression of only convolutional layers. (b) compression of whole layers including fully-connected layers. Smaller error is better.}
\vspace{-4.5ex}
\label{fig:comp_new}
\end{figure}

\textbf{AlexNet with ImageNet.}\;\; In \cite{ydkim}, the authors show the compression results for FLOPs and parameters. 
It is not possible to use ENC-Map under both these constraints as it provides mapping only from one of these constraints to the accuracy. However, ENC-Model and ENC-Inf can be used by populating $\mathbf{R}$ by rank configurations that satisfy both the FLOPs and parameters constraints. We have used $N$=50 here and optimize both convolutional and fully-connected layers.
Initial learning rate of $10^{-3}$ was used, which was reduced by a factor of 2 after every 2 epochs till the 16-th. Fine-tuning takes around 9 hours. As denoted in Table~\ref{table:cost_all}, our method shows 1.8\% higher top-5 accuracy than \cite{ydkim} at the same complexity. To find the rank configuration, our algorithm only takes only a minute with a single core CPU with ENC-Model and 3 minutes on 4 GPUs with ENC-Inf. 
Fig.~\ref{fig:comp_new} indicates that our method outperforms the state-of-the-art.
The FLOPs saving ratios of reffed works in Fig.~\ref{fig:comp_new} are calculated in our experiment.
We achieve 43.33\% and 42.40\% top-1 error at 30.7\% and 95.0\% FLOPs for Fig.~\ref{fig:comp_new}(a). Also, we attain the 20.4\% top-5 error at 30\% FLOPs for Fig.~\ref{fig:comp_new}(b).
Compared to \cite{lrd,lra} using the accuracy compensation techniques such as knowledge transfer~\cite{lrd} and force regularization~\cite{lra}, our method uses only fine-tuning which is the most simple and effective strategy.

\begin{table}[t]
\vspace{-1.5ex}
\centering
\begin{tabular}{ccc}
\toprule
\begin{tabular}[c]{@{}c@{}}VGG-16\\ (FLOPs 20\%)\end{tabular} &
\begin{tabular}[c]{@{}c@{}}Top-5\\$\Delta$Acc.\end{tabular} &
\begin{tabular}[c]{@{}c@{}}Search Range\end{tabular}            \\\midrule

Asym.3D~\cite{asym}& \begin{tabular}[c]{@{}c@{}}1.0\%\end{tabular} & 
\multirow{3}{*}{\begin{tabular}[c]{@{}c@{}}Layer-by-layer\end{tabular}}\\

Y-D \textit{et al}.{~\cite{ydkim}}& \begin{tabular}[c]{@{}c@{}}0.5\%\end{tabular}& \\

CP-3C{~\cite{cpd}}& \begin{tabular}[c]{@{}c@{}}0.3\%\end{tabular}& \\
\midrule
\begin{tabular}[c]{@{}c@{}}\textbf{ENC-Inf} \end{tabular}& 
\begin{tabular}[c]{@{}c@{}}\textbf{0.0}\% \end{tabular}& 

\begin{tabular}[c]{@{}c@{}}Whole-network\end{tabular}   
\\ \bottomrule
\end{tabular}
\caption{Comparison over target complexity of 20\% FLOPs with VGG-16. Baseline top-5 accuracy is 89.9\%.}
\vspace{-3ex}
\label{table:vgg_20}
\end{table}

\textbf{VGG-16 with ImageNet.}\;\; We optimize the convolutional layers under a FLOPs constraint.
The complexity constraint is set to 25\% and 20\% FLOPs to compare with \cite{adc,ars} and \cite{asym,ydkim,cpd}, respectively. 
For ENC-Inf, we set $N$=40. 
The layers 7-8, 9-10, and 11-13 are grouped each other for top-level sub-spaces.
The compressed network was fine-tuned with an initial learning rate of $10^{-5}$, which was decreased by a factor of 10 at the fourth epoch.
Fine-tuning takes about 1 day.
With 25\% FLOPs reduction, ENC-Inf takes only 5 minutes at 4 GPUs, and it shows the 2.3\% and 0.3\% higher top-1 accuracy without and with fine-tuning, respectively, compared to~\cite{ars}. 
As denoted in Table~\ref{table:cost_all}, our ENC-Map is extremely fast. It can find the result in only 4 seconds with a single core CPU, while the previous research takes 4 hours at 8 GPUs~\cite{adc} and 7 hours at 4 GPUs~\cite{ars}.
Results with 20\% FLOPs reduction are given in Table~\ref{table:vgg_20}. It is seen that our method outperforms the layer-by-layer strategies. 

\textbf{ResNet-56 with Cifar10.}\;\; 
The search space is defined in 3-level hierarchy.
The layers 2-19, 21-37, and 39-55 are placed in separated groups excluding first convolutional layer and fully-connected layer.
We set the the maximum number of bottom layers in a group as four.
The second-level spaces are defined by the number of bottom layers such as Fig.~\ref{fig:group}.
For ENC-Inf, we set $N$=20.
The compressed ResNet-56 is fine-tuned with an initial learning rate of $10^{-3}$, which is reduced by a factor of 10 at the 16th, 24th and 32nd epoch. Fine-tuning takes around 1 hour. ENC-Map and ENC-Model achieve a 0.1\% accuracy loss (i.e. 93.1\% top-1 accuracy) with 50\% FLOPs reduction after fine-tuning. The times taken by ENC-Map and ENC-Inf are 3 seconds with single core CPU and 5 minutes with 4 GPUs, respectively. 
It is much more efficient compared to the learning based state-of-the-art, which takes 1 hour on a single GPU~\cite{amc}.


\begin{table}[t]
\centering
\begin{tabular}{cccc}
\toprule
\begin{tabular}[c]{@{}c@{}}Model\end{tabular} & \begin{tabular}[c]{@{}c@{}}Searching\\Policy\end{tabular} & FLOPs & \begin{tabular}[c]{@{}c@{}}Top-1\\ $\Delta$Acc.\end{tabular}    \\ \midrule
AlexNet & ENC-Inf& 31\%  & 0.0\%  \\ 

VGG-16 & ENC-Model & 24\%  & -0.4\%  

\\ 
ResNet-56& ENC-Map & 55\%  &{-0.1}\%
\\ \bottomrule
\end{tabular}
\caption{Lossless compression with full fine-tuning.}
\vspace{-3.0ex}
\label{table:lossless}
\end{table}

\begin{table}[t]
\centering
\begin{tabular}{cccc}
\toprule
\begin{tabular}[c]{@{}c@{}}Searching\\Policy\end{tabular} & \begin{tabular}[c]{@{}c@{}}Fine-tuning\\ Epochs\end{tabular} & FLOPs & \begin{tabular}[c]{@{}c@{}}Top-1\\ $\Delta$Acc.\end{tabular}    \\ \midrule
ADC~\cite{adc}& 0& 64\%  & 0.0\%  \\
\textbf{ENC-Model}& \textbf{0}& \textbf{57}\%  & \textbf{-0.1}\% \\ \midrule
\multirow{3}{*}{
\textbf{ENC-Model}}
& 0.1& 41\%  & -0.1\% \\
& 0.2& 39\%  & -0.1\% \\
& \textbf{1}& \textbf{33}\%  & \textbf{0.0}\%  
\\ \bottomrule
\end{tabular}
\caption{Fast lossless compression with brief fine-tuning for VGG-16. Baseline top-1 accuracy is 70.6\%. Smaller $\Delta$Acc. is better.}

\vspace{-3ex}
\label{table:diff_epoch}
\end{table}

\subsection{Lossless Compression}
The idea of lossless compression is to set the target accuracy to that of the original neural network and then optimize the network. However, since we use fine-tuning after optimization, we can set the target accuracy to slightly lower than that of the original neural network. The remaining accuracy is recovered by fine-tuning. 
To set the reduced accuracy constraint, we use the accuracy estimation method in \cite{ars}.
Then, the complexity constraint (i.e. FLOPs) is calculated from the mapping function between the accuracy and complexity constraints.
As summarized in Table~\ref{table:lossless}, our compression methods achieve a significant reduction in FLOPs without accuracy loss.

The combination of our fast compression method and brief fine-tuning can further reduce the FLOPs without accuracy loss.
Table~\ref{table:diff_epoch} shows the result of lossless compression with fine-tuning under 1 epoch.
The accuracy thresholds considering fine-tuning are calculated for 0.1, 0.2, and 1 epochs using the method in~\cite{ars}.
We reduced the FLOPs by 41\% with VGG-16 without any accuracy loss. The process took only 0.1 epoch or 22 minutes with a single GPU. 
Also, our method with 1 epoch fine-tuning takes 3.7 hours with a single GPU, and it provides better compression (33\% FLOPs) compared to the 4 hour search (64\% FLOPs) in~\cite{adc}.

\subsection{Results on an Embedded Board}
\begin{figure}[t]
\begin{center}
\includegraphics[width=0.99\linewidth]{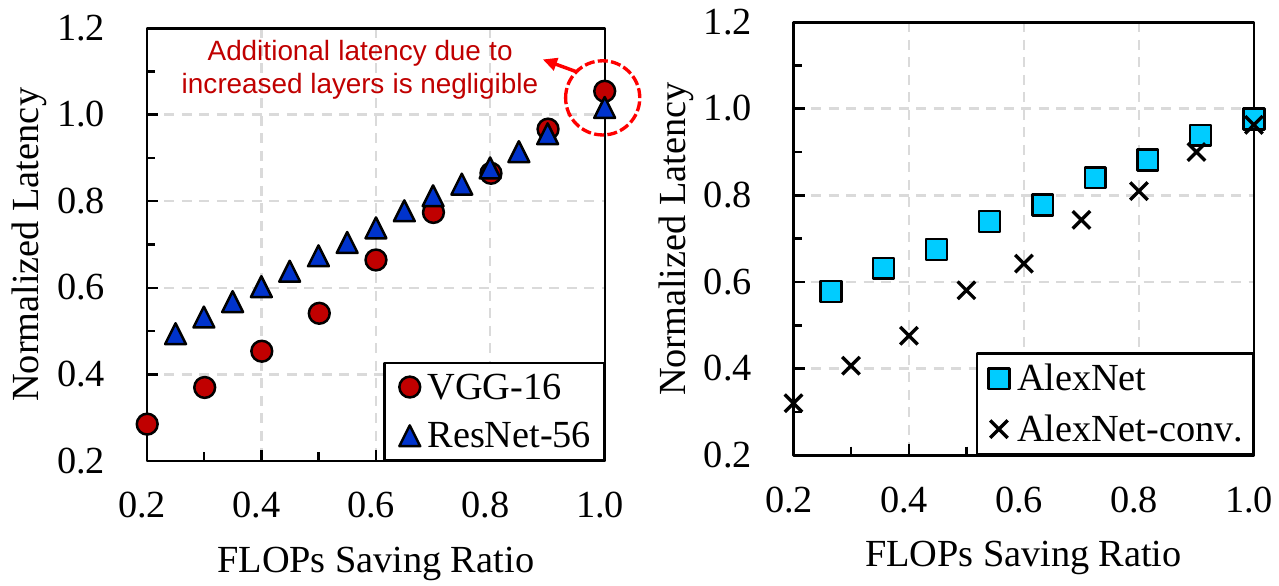}
\end{center}
\vspace{-1ex}
\caption{Normalized CPU latency of optimized models. The latency of baseline CNN is 10.59s in VGG-16, 142.2ms in ResNet-56, and 825.0ms in AlexNet for single image classification.}
\vspace{-3.0ex}
\label{fig:speed}
\end{figure}

We evaluate the latency of the compressed network models for inference on the ODROID-XU4 board with Samsung Exynos5422 mobile processor.
Fig.~\ref{fig:speed} shows that the FLOPs saving ratio in our paper is directly related to the real latency for single image inference. 
We use ENC-Model for optimizing the convolutional layers of AlexNet, VGG-16, and ResNet-56 in Fig.~\ref{fig:speed}.
We note that the latency of convolutional layers is 72\% and 62\% of total latency for the baseline ResNet-56 and AlexNet, respectively, and the bias latency in Fig.~\ref{fig:speed} is due to the other operations including batch normalization and fully-connected layers.
The results of AlexNet-conv indicate that the theoretical FLOPs reduction of convolutional layers corresponds to the latency improvement of those layers.

\section{Conclusion}
In this paper, we propose the efficient neural network compression methods. Our methods are based on low-rank kernel decomposition.
We propose a holistic, model-based approach to obtain the rank configuration that satisfies the given set of constraints.
Our method can compress the neural network while providing competitive accuracy.
Moreover, the time taken by our method for compression is in seconds or minutes, whereas previously proposed methods take hours to achieve similar results.
\\

\noindent \textbf{Acknowledgement}\;\; 
This work was supported by MSIT as GFP / (CISS-2013M3A6A6073718).


{\small
\bibliographystyle{ieee}


}

\end{document}